\documentclass[nohyperref]{article}

\usepackage{microtype}
\usepackage{graphicx}
\usepackage{subfigure}
\usepackage{booktabs} 
\usepackage{multirow}
\usepackage{hyperref}

\usepackage{amsmath}
\usepackage{setspace}
\usepackage{mathrsfs}
\usepackage{amsfonts}

\usepackage[accepted]{icml2022}

\usepackage{amsmath}
\usepackage{amssymb}
\usepackage{mathtools}
\usepackage{amsthm}

\newcommand{\x}{\mathbf{x}}

\usepackage{amsmath}

\DeclareMathOperator*{\argmin}{arg\,min}

\usepackage[capitalize,noabbrev]{cleveref}

\theoremstyle{plain}

\theoremstyle{definition}

\theoremstyle{remark}

\usepackage[textsize=tiny]{todonotes}

\icmltitlerunning{Submission and Formatting Instructions for ICML 2022}

\begin{document}

\twocolumn[
\icmltitle{High Performance of Gradient Boosting in Binding Affinity Prediction}



\icmlsetsymbol{equal}{*}

\begin{icmlauthorlist}
\icmlauthor{Dmitrii Gavrilev}{skoltech}
\icmlauthor{Nurlybek Amangeldiuly}{skoltech}
\icmlauthor{Sergei Ivanov}{criteo,skoltech}
\icmlauthor{Evgeny Burnaev}{skoltech}
\end{icmlauthorlist}

\icmlaffiliation{skoltech}{Skolkovo Institute of Science and Technology}
\icmlaffiliation{criteo}{Criteo AI}

\icmlcorrespondingauthor{Dmitrii Gavrilev}{Dmitrii.Gavrilev@skoltech.edu}

\icmlkeywords{Machine Learning, ICML}

\vskip 0.3in
]



\printAffiliationsAndNotice{\icmlEqualContribution} 

\begin{abstract}
Prediction of protein-ligand (PL) binding affinity remains the key to drug discovery. Popular approaches in recent years involve graph neural networks (GNNs), which are used to learn the topology and geometry of PL complexes. However, GNNs are computationally heavy and have poor scalability to graph sizes. On the other hand, traditional machine learning (ML) approaches, such as gradient-boosted decision trees (GBDTs), are lightweight yet extremely efficient for tabular data. We propose to use PL interaction features along with PL graph-level features in GBDT. We show that this combination outperforms the existing solutions.
\end{abstract}

\section{Introduction}
Binding affinity is one of the major determinants to identify whether a molecule binds to a desired macromolecular target. It helps speculate if the molecule can induce a biological response in an organism, which is the central mechanism in drug action. Reliable and fast prediction of such parameter is an important and challenging task in cheminformatics \cite{shunzhou2020}. To build an accurate binding affinity prediction model PL complex 3D structures are needed \cite{jongtae2021}, and are readily accessible from steadily growing open source databases \cite{alphafold2021}. While docking is an established approach to predict affinity \cite{kitchen2004}, recent ML-based algorithms consistently outperform existing classical methods \cite{meli2021}. Considered a natural way of processing chemical structures, recently GNNs are proposed to predict binding affinity. In a typical prediction pipeline for cheminformatics, a GNN generalizes to a labeled and weighted graph, nodes of which represent molecular atoms, and edges may represent their bonds by including various interaction features such as bond angles and distances \cite{jones2021, jongtae2021}. This application domain of GNN can lead to a promising results \cite{Kearnes2016}. The main bottleneck of GNNs is, however, their computationally complex iterative convergence \cite{gupta2020hybrid, tiezzi2020lagrangian}. 

In contrast, GBDTs provide faster convergence and perform out-of-the-box, which makes them a popular tool for ML contests and industrial applications \cite{ivanov2021, Gorishniy2021}. Notwithstanding the advantages, powerful model of GBDTs often overlooked \cite{50030}.

In this work, we leverage the advantages of GBDTs to graph-level structural and interaction feature representations of PL complexes for binding affinity prediction. Assessment of the model performance on external test sets showed that the GBDTs can perform equally well or outperform the recent state-of-the-art GNN-based approach, SIGN \cite{sign}.
\begin{table*}[t]
    \caption{Metrics on PDBbind core and CSAR-HiQ sets.}
    \label{tab:metrics}
    \footnotesize
    \centering
    \begin{tabular}{@{}lcccccccc@{}}
        \toprule
        \multirow{2}{*}[-1em]{Models} & \multicolumn{4}{c|}{PDBbind core set} & \multicolumn{4}{c}{CSAR-HiQ set}\\
        \midrule
        {} & RMSE $\downarrow$ & MAE $\downarrow$ & SD $\downarrow$ & R $\uparrow$ & RMSE $\downarrow$ & MAE $\downarrow$ & SD $\downarrow$ & R $\uparrow$\\
        \midrule
        LR & 1.608 (0.000) & 1.285 (0.000) & 1.576 (0.000) & 0.690 (0.000) & 1.994 (0.000) & 1.555 (0.000) & 1.909 (0.000) & 0.650 (0.000) \\
        
        MLP & 1.456 (0.032) & 1.158 (0.025) & 1.454 (0.033) & 0.744 (0.014) & 2.195 (0.203) & 1.727 (0.178) & 2.046 (0.116) & 0.575 (0.066) \\

        SIGN & 1.316 (0.031) & 1.027 (0.025) &  1.312 (0.035) & 0.797 (0.012) & 1.735 (0.031) & 1.327 (0.040) & 1.709 (0.044) & 0.754 (0.014) \\
        \hline
        \textbf{Ours: CatBoost} & 1.321 (0.008) & 1.045 (0.011) & 1.270 (0.012) & .812 (0.004) & 1.798 (0.031) & 1.391 (0.040) & 1.679 (0.014) & 0.744 (0.005) \\
        
        \textbf{Ours: LightGBM} & \textbf{1.316 (0.010)} & 1.040 (0.007) & 1.279 (0.015) & 0.809 (0.005) & \textbf{1.725 (0.038)} & 1.305 (0.046) & 1.660 (0.049) & 0.751 (0.017) \\
        \bottomrule
    \end{tabular}
\end{table*}

\section{Related Work}
The early ML methods of binding affinity prediction rely on careful feature engineering. Namely, RF-Score proposed training a Random Forest model on intermolecular interaction features \cite{mlfeatures}. For sets of atom types $P$ in protein and $L$ in ligand, the authors count the number of atom type pairs in the complex. More formally, the resulting interaction features vector $x$ is $|P| \times |L|$ dimensional and 
\begin{equation*}
    x_{ji} = \sum_{a \in protein} \sum_{b \in ligand} \delta_{aj} \delta_{bi} \Theta \left(d_{cutoff} - d_{ab} \right),
\end{equation*}
where $\delta$ and $\Theta$ are the Kronecker and Heavyside functions respectively, $d_{cutoff}$ is the interaction cutoff distance and $d_{ab}$ is the distance between atoms $a$ and $b$. Particularly, the authors set $d_{cutoff} = 12$. As for $P$ and $L$, the following sets are recommended:
\begin{gather*}
    P = \left\{ C, N, O, S \right\}, \\
    L = \left\{ C, N, O, F, P, S, Cl, Br, I \right\}.
\end{gather*}
Therefore, the interaction features can be described by a 36-dimensional vector. Designing representations for the molecules has also received a great deal of attention in the context of graph kernels~\cite{kriege2020survey, ivanov2018anonymous} which have been applied successfully for various applications on graphs~\cite{sharaev2018learning, ivanov2018learning}. 
On the other hand, recent trends in binding affinity prediction involve extensive usage of GNNs to obtain spatial representations. 
Such approach also allows characterizing intermolecular and intramolecular interactions. 
To enrich the representation, the efforts are being directed towards integrating atomic distances \cite{lim2019}, bond angles and physics-derived descriptors \cite{moon2022pignet}. 
In such work \cite{jiang2021interactiongraphnet}, authors construct protein graphs, ligand graphs and PL graphs separately to represent PL structure and interaction by also integrating distance, angle, and various 2D node and edge features. 
Another approach, graphDelta \cite{karlov2020graphdelta} characterizes binding site of the complex by computing Behler-Parrinello symmetry functions \cite{bps} per each common PL heavy atom types in the complex and obtains complex interaction features to train a message-passing neural network.
Very recent work, SIGN exploits the interaction features similar to RF-Score.
Moreover, the authors of SIGN propose to use not only topological information but also angles and distances. 
Despite outperforming ML and NN-based models, following drawbacks inherent to GNN models are not solved in SIGN:
\begin{itemize}
  \item Time-consuming construction of graphs from raw 3D coordinates \cite{wang2012fast};
  \item Requires large computational resources for training \cite{liu2020efficient}.
\end{itemize}
We provide time and memory consumption details of SIGN in Section \ref{par:training}.
\section{Our Approach}
\label{sec:approach}
In this work, we incorporate PL interaction features, as introduced in RF-Score, alongside graph-level features. To construct graph-level features, the atomic features are pooled together. For each atomic feature, we sum the values over all atoms in the complex. Additionally, we calculate the standard deviation of atomic features to characterize atomic distribution statistics. Categorical features were encoded to a one-hot numeric array: atom types were transformed into a binary array, where each bit encodes a type. Then, the graph-level feature constructed from these arrays represents the distribution of atom types in a complex. Below we list the features that we used to train our models:
\begin{itemize}
  \item RF-Score PL interaction features;
  \item Summarized graph-level atomic features;
  \item Standard deviation of atomic features;
  \end{itemize}
After obtaining features, we pass them to a GBDT model. GBDT \cite{friedman2001greedy} is a widely used algorithm that builds ensemble of decision trees. The idea of GBDT is to iteratively add weak decision trees and aggregate their predictions to obtain an overall strong model. At each iteration the model is updated by building a new tree that minimizes the loss function w.r.t. previous predictions of the ensemble. In particular, at each iteration $t$ the model $f(\x)$ is updated in an additive manner:
\begin{equation}
\label{eq:additive}
f^{t}(\x) = f^{t-1}(\x) + \epsilon\, h^{t}(\x),
\end{equation}
where $f^{t-1}$ is previously built trees, $h^{t}$ is a weak learner that is chosen from some family of functions $\mathcal{H}$ (e.g. decision trees), and $\epsilon$ is a learning rate. The weak learner $h^{t}\in \mathcal{H}$ approximates the direction of the negative gradient of a loss function $L$ w.r.t.~the current model's predictions:
\begin{equation}
\label{eq:tree}
h^{t} = \argmin_{h\in \mathcal{H}} \sum_i \left(-\frac{\partial L(f^{t-1}(\x_i), y_i)}{\partial f^{t-1}(\x_i)}- h(\x_i)\right)^2 . 
\end{equation}
As such one can think of GBDT's training as functional gradient descent, where at each iteration we move in the direction that minimizes the loss. This allows flexibly training GBDT on various tabular continuous/discrete, noisy and heterogeneous data.

\section{Experiments}
\label{sec:experiments}
\paragraph{Data.}
For evaluating the performance of the models, we use the same dataset and splits as in \cite{sign}. Specifically, all experiments were conducted on PDBbind v2016 \cite{pdbbind} and CSAR-HiQ \cite{csar}, which contain experimental PL binding affinity data. The core set of PDBbind (290 complexes) was used as a benchmark for model comparison \cite{core}. The models were trained on the difference between the refined and core sets (3767 complexes). For tuning hyperparameters, 377 out of 3767 complexes were used as a validation set. After training and tuning, the models were additionally evaluated on CSAR-HiQ dataset. Since CSAR-HiQ and the refined set of PDBbind overlap, we filter out these overlapping complexes for evaluation. The filtered dataset consists of 135 complexes.

\begin{figure}[t]
\vskip 0.2in
\begin{center}
\centerline{\includegraphics[width=\columnwidth]{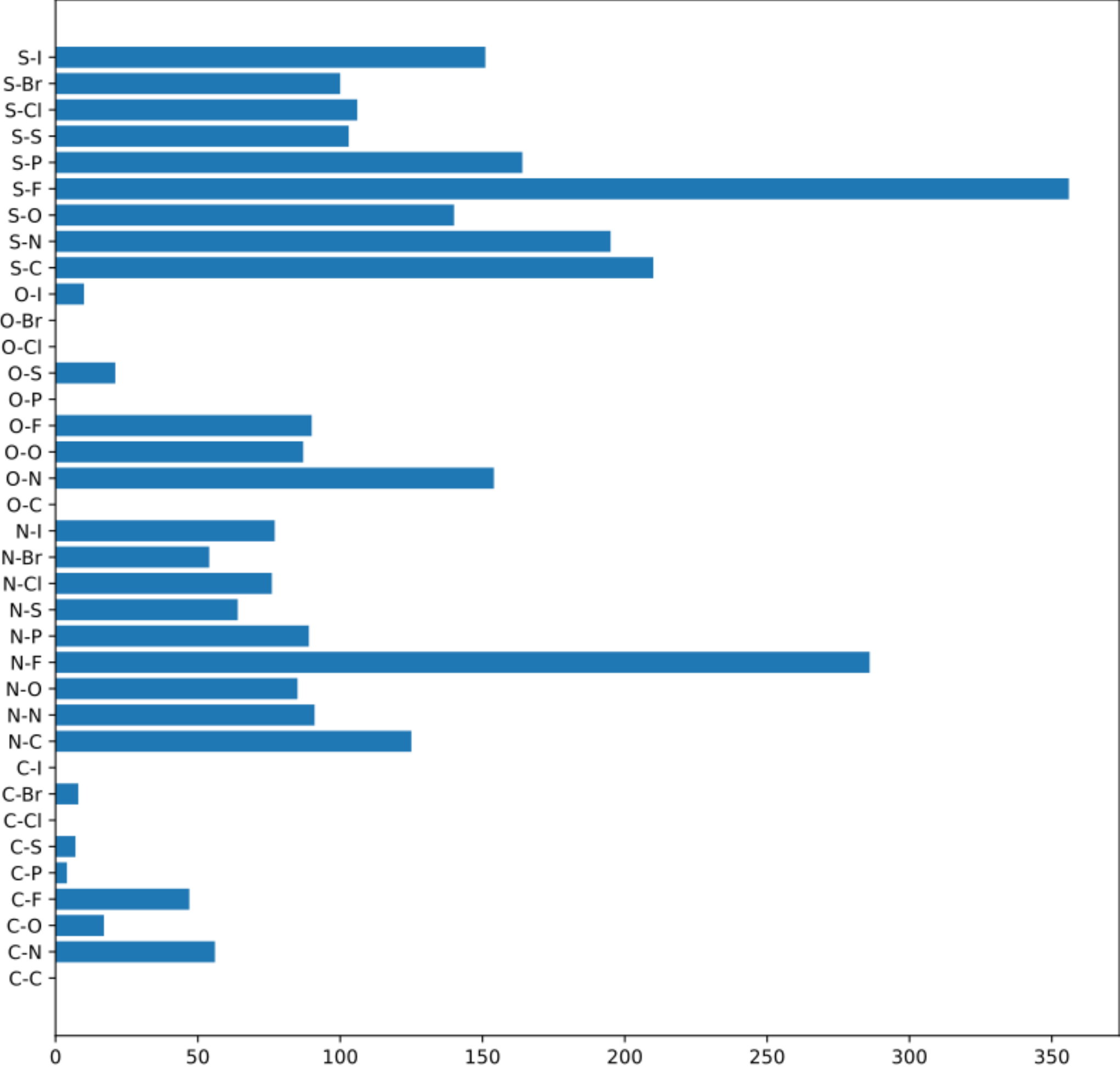}}
\caption{Feature importance (interaction features).}
\label{fig:inter_importances}
\end{center}
\vskip -0.2in
\end{figure}

We follow the same feature extraction procedure as the authors of SIGN. Each atom is represented by a 36-dimensional vector, 18 of which describe protein features and the rest describe ligand features. Those features include atom type, hybridization, partial charge, aromaticity, etc. For this set of features, we run the pipeline presented in Section \ref{sec:approach}. As a result, we obtain 108 features for each complex — 36 from the interaction matrix, 36 by summing atomic features, and the other 36 by calculating standard deviation.

\paragraph{Models.}
Our work focuses mainly on GBDT models since they have shown great performance for tabular data in both time and quality. We compare two popular GBDT models, CatBoost \cite{catboost} and LightGBM \cite{lightgbm}, with SIGN.
First, we train the models on PDBbind refined set and tune their hyperparameters on validation set. For tuning, we utilize Optuna \cite{optuna}, automatic hyperparameter search framework. While validating, we aim to minimize root mean squared error (RMSE). The best hyperparameters for CatBoost are following: depth = 10, grow policy = symmetric tree, L2 leaf regularization = 73.47. As for LightGBM, the optimal hyperparameters are: bagging fraction = 1, bagging frequency = 0, feature fraction = 0.8, L1 regularization = 0, L2 regularization = 0, minimal child samples = 20, number of leaves = 244.
\begin{figure}[t]
\vskip 0.2in
\begin{center}
\centerline{\includegraphics[width=\columnwidth]{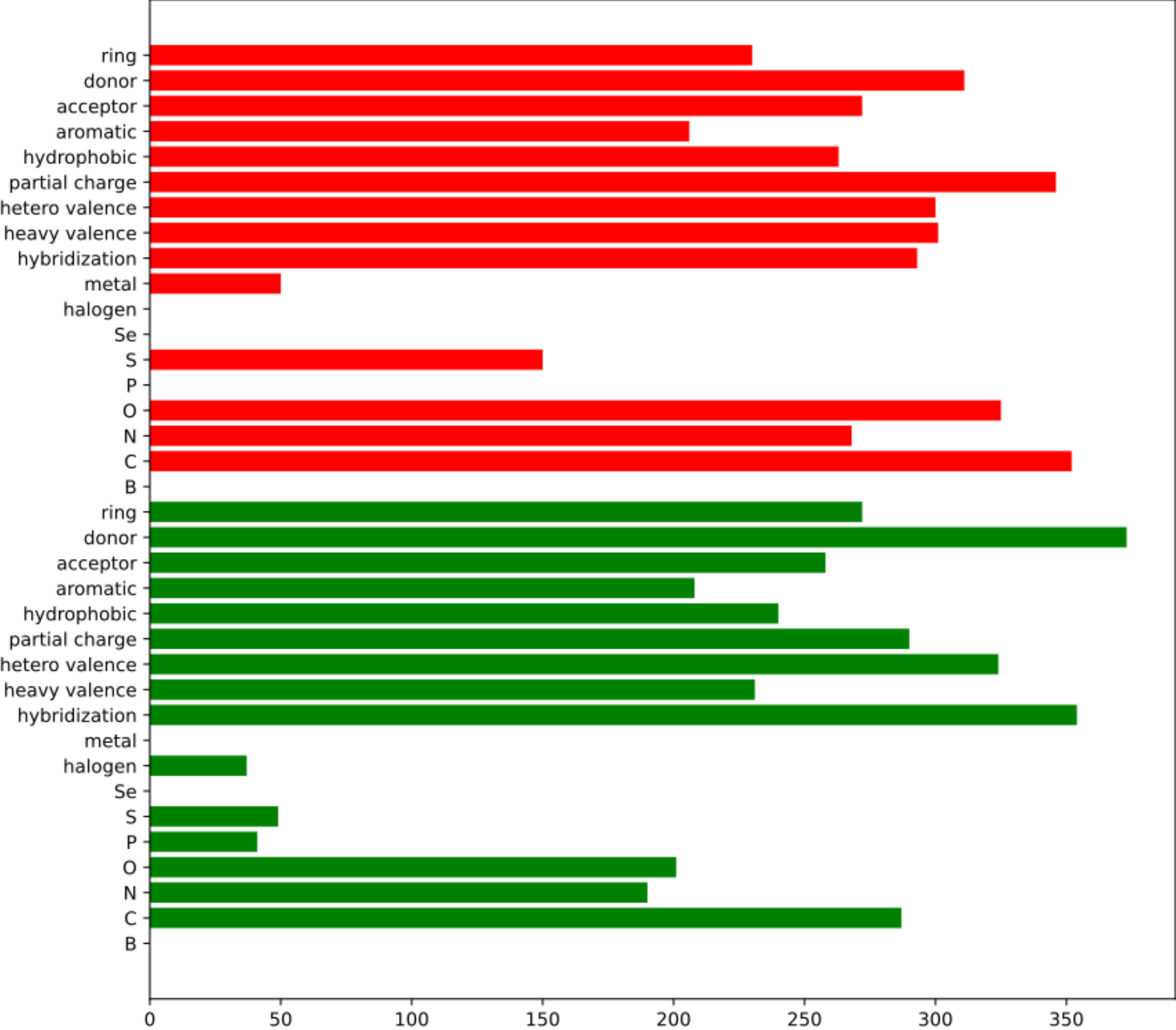}}
\caption{Feature importance (sum of atomic features).}
\label{fig:sum_importances}
\end{center}
\vskip -0.2in
\end{figure}

\begin{figure}[t]
\vskip 0.2in
\begin{center}
\centerline{\includegraphics[width=\columnwidth]{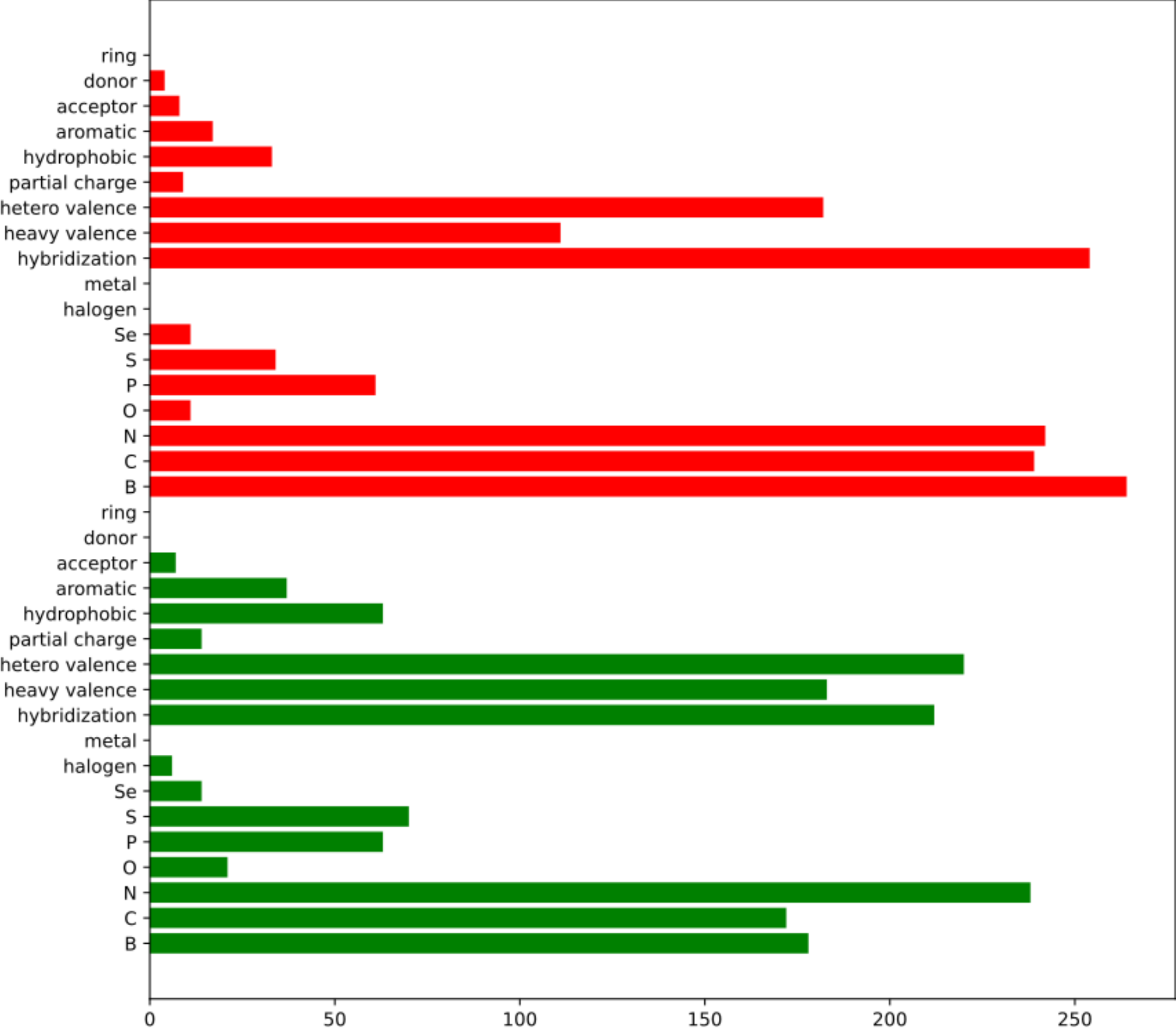}}
\caption{Feature importance (standard deviation of atomic features).}
\label{fig:std_importances}
\end{center}
\vskip -0.2in
\end{figure}

Next, we evaluate the models on the PDBbind core set. Table \ref{tab:metrics} summarizes our main findings. The following metrics are reported: RMSE (root mean squared error), MAE (mean absolute error), SD (standard deviation), R (Pearson's correlation). The metrics are averaged across 5 different runs (their standard deviation are presented in parentheses). The best values are denoted in bold. On the core set, LightGBM has the same RMSE as SIGN on average but it deviates less, while CatBoost performs better in terms of SD and R. The only metric on which SIGN outperforms GBDT model is MAE. Besides GBDT models, we also provide the results for linear regression and three-layer perceptron.

To assess the generalization ability, the models are further evaluated on CSAR-HiQ set. As can be seen from Table \ref{tab:metrics}, classical GBDT models perform on par with SIGN method. LightGBM demonstrates slight improvement over SIGN in terms of RMSE and is able to outperform all presented models. Nevertheless, SIGN shows better R score. It should be noted that GBDTs achieved these results easily, owing to efficiency and scalability of their architecture. 
\paragraph{Feature importance.}
Finally, we obtain the feature importance to analyze their contribution. Figures \ref{fig:sum_importances} and \ref{fig:std_importances} show the degree of contribution of atomic features (their sum and standard deviation). Green and red bars correspond to ligand and protein features respectively. Figure \ref{fig:inter_importances} illustrates the importance of interaction features. It is notable that mainly uncommon interactions in the complexes, such as $S-F$, $N-F$ have the highest importance in decision making of the model. It is also important to note that the sum of atomic features has a more even distribution of importance than those of interaction features while having the same peak in magnitude. This might suggest that pooling atomic-level features into graph-level features can significantly boost the performance of RF-Score.  
\paragraph{Training time and memory consumption.}
\label{par:training}
We also report the benefits of using GBDTs in terms of speed and memory consumption. All experiments were conducted on 22 CPU cores and 1 NVIDIA A10 GPU (GPU was used for training SIGN). While training SIGN took 11 hours (300 epochs with batch size set to 16), training CatBoost took only 3 minutes. Moreover, LightGBM has been trained only under 4 seconds in total. As for memory consumption, SIGN allocates 19 GB of RAM and 13 GB of VRAM, whereas GBDT models required less than 1 GB of RAM.

\section{Discussion and Future Work}
In this work, we looked back at traditional ML methods for predicting PL binding affinity and proposed a way to improve their performance. We showed that the inclusion of graph-level features with their standard deviation results in better metrics. To our surprise, GBDT models can be comparable to heavy GNN models without much effort. We hope that our findings would bring attention back to simpler models, as there might be more undiscovered potential. 
Next, we plan to integrate GBDT and GNN into one end-to-end architecture to utilize the advantages of heterogeneous learning and representation learning, as recently reported in our work, BGNN \cite{ivanov2021}. The approach demonstrated state-of-the-art performance and efficiency results on node-level prediction tasks, and adapting the architecture to process the graph-level molecular features is a promising direction.

\bibliography{example_paper}
\bibliographystyle{icml2022}


\end{document}